\documentclass{article}
\usepackage{spconf,amsmath,graphicx,hyperref}
\usepackage{booktabs}
\usepackage{marvosym} 

\title{Remember by Asking: Retrieval-Induced Memory Evolution \\
for LLM Agents}
\name{%
Wanqi Zhou$^{1,*}$, Jiawei Lu$^{1,*}$, Yang Wang$^{2}$,
Zhaolong Xing$^{1}$, Zhen Chen$^{1}$, Ai Han$^{1}$,
Haoyue Shi\textsuperscript{2,\Letter}%
\thanks{\Letter\ Corresponding author: haoyueshi@chd.edu.cn}
\thanks{$^{*}$ These authors contributed equally to this research.}
}
\address{$^{1}$JD.com \qquad $^{2}$Chang'an University}
\begin{document}
%
\maketitle
\begin{abstract}
Long-term memory is essential for language agents to maintain coherent and effective behavior over extended, multi-session interactions. 
Existing memory systems mainly use retrieval at read time, while write-time memory formation still relies on direct extraction or compression.
However, when future information needs are unknown, compressing an entire interaction in one pass can overlook locally important details that may matter later.
To this end, we introduce \textbf{RIME}, a retrieval-induced memory framework that shifts memory construction from monolithic compression toward evidence-centered integration. 
RIME uses generic self-questions to retrieve focused dialogue evidence and grounds memory formation in both the retrieved evidence and relevant historical memories, which are jointly reconciled into an evolving memory bank with temporal and provenance information. 
At inference time, compressed memory serves as the primary rather than the sole source of evidence: when it cannot support an answer, RIME retrieves relevant source dialogue together with its local context to recover information omitted during memory formation, without resorting to full-history processing.
Extensive experiments on LoCoMo with Qwen3-235B-A22B and GPT-5.6 Sol show that RIME consistently achieves the best performance across all three quality metrics among the compared methods, while requiring substantially fewer query-time LLM tokens.

\end{abstract}
\section{Introduction}
\label{sec:intro}

Memory is fundamental for language agents operating over extended interactions. 
Without persistent memory, agents cannot reliably retain user-specific information, connect events across sessions, or reuse past experience in future decisions. 
Prior work has therefore developed mechanisms for storing, retrieving, and managing information beyond the immediate context window, ranging from memory streams and memory banks to more structured and adaptive memory systems~\cite{park2023generative,zhong2024memorybank,packer2023memgpt,chhikara2025mem0,xu2026mem,ma2026deserves}.

Despite this progress, most systems still form memories by directly extracting or compressing conversational context. 
This can overlook useful evidence, while details discarded at formation time may later become important for unforeseen queries. 
Relying only on compressed memory prevents such information from being recovered, whereas full-history fallback incurs substantial inference cost~\cite{you2026dmem}. 
These limitations motivate selective access to source dialogue during both memory formation and memory use.

\begin{figure*}[t]
    \centering
    \includegraphics[width=0.91\textwidth]{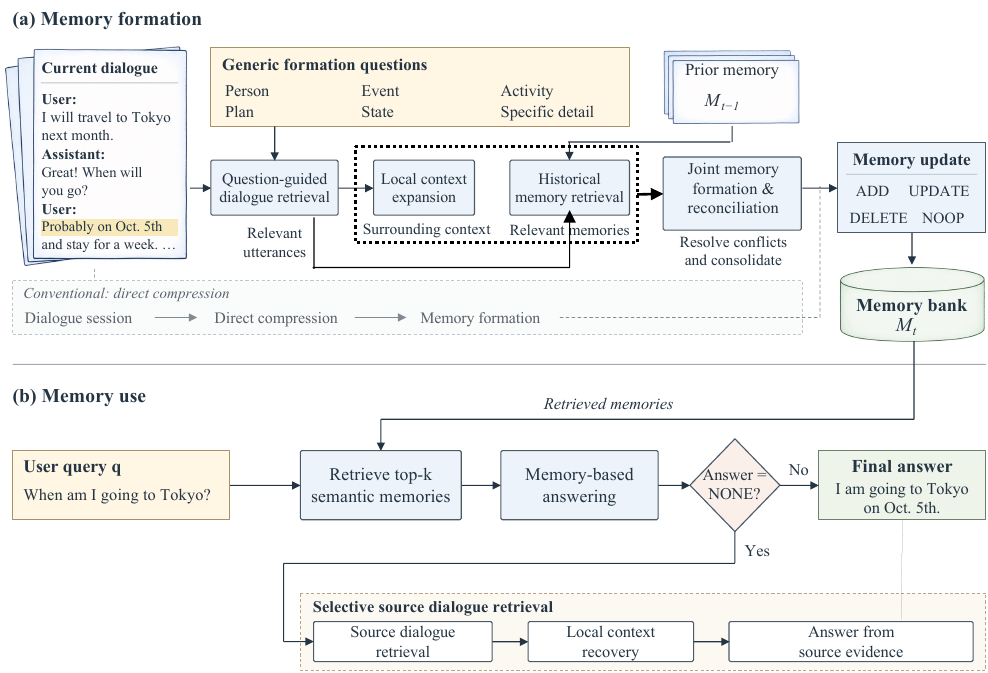}
    \caption{
    Overview of RIME.
    (a) During memory formation, generic formation questions are used as retrieval cues to identify focused dialogue evidence, which is expanded with local context and combined with relevant historical memories for joint memory formation and reconciliation.
    Unlike conventional direct compression, RIME retrieves relevant evidence before consolidation.
    (b) During memory use, RIME first answers from retrieved semantic memories and selectively returns to the source dialogue only when the memory-based answer is insufficient.
    }
    \label{fig:rime_framework}
\end{figure*}

To this end, we introduce \textbf{RIME}, a retrieval-induced framework for long-term dialogue memory. 
As illustrated in Fig.~\ref{fig:rime_framework}, rather than compressing an entire interaction at once, RIME uses generic self-questions to retrieve focused dialogue evidence, which is combined with local context and relevant historical memories to incrementally form an evolving memory bank. 
By grounding memory formation in retrieved evidence, RIME surfaces potentially important information before consolidation, reducing the risk that it is overlooked in the full conversational context.
At inference time, RIME treats compressed memory as the primary rather than the sole information source, selectively retrieving relevant source dialogue and surrounding context when needed instead of processing the full interaction history.
Across LoCoMo experiments with Qwen3-235B-A22B \cite{yang2025qwen3} and GPT-5.6 Sol \cite{openai2026gpt56}, RIME consistently improves both memory effectiveness and inference efficiency over existing memory systems.

Our contributions are threefold. (1) We introduce a question-guided retrieve-then-consolidate mechanism for memory formation, where generic self-questions first retrieve focused dialogue evidence before memory construction, rather than relying on a single global compression of the full interaction. (2) We introduce selective source-context recovery for memory use, allowing compressed memory to serve as the primary but not exclusive information source and recovering relevant dialogue context only when additional evidence is needed. (3) We conduct extensive experiments on LoCoMo with Qwen3-235B-A22B and GPT-5.6 Sol, where RIME achieves the strongest performance across LLM-Judge, F1, and BLEU-1 among the compared methods while requiring substantially fewer query-time LLM tokens than deliberative memory baselines.

\section{Related Work}
\label{sec:related_work}

\noindent\textbf{Long-term memory for language agents.}
Long-term memory systems have evolved from persistent external memory
stores~\cite{park2023generative,zhong2024memorybank,packer2023memgpt} toward
more sophisticated mechanisms for constructing and maintaining compact memory.
Mem0~\cite{chhikara2025mem0} consolidates extracted facts through explicit
memory operations, while A-MEM~\cite{xu2026mem} organizes memories as
dynamically linked and evolving notes.
Nemori~\cite{ma2026deserves} and FadeMem~\cite{wei2026fademem}
focus on selective retention, through semantic distillation and adaptive
forgetting, respectively.
SimpleMem~\cite{simplemem2026} and LightMem~\cite{fang2026lightmem} improve
efficiency through structured compression, consolidation, and hierarchical
memory management.
Despite their different designs, these methods mainly optimize how observed
interaction content is transformed, organized, or maintained as memory.
RIME intervenes one step earlier: through generic self-questions, it brings
retrieval into the memory-writing process, recalling dialogue evidence relevant
to each question before consolidation to better preserve important information
that may otherwise be overlooked.

\noindent\textbf{Retrieval and memory-based reasoning.}
Beyond memory construction, recent work has explored richer mechanisms for
accessing and reasoning over long-term memory.
Hindsight~\cite{latimer2026hindsight} combines multiple retrieval strategies
with explicit reflection over structured memories, while
SAGE~\cite{wang2026sage} and REALM~\cite{song2026retrieval} use retrieval feedback
to further evolve or reconsolidate the memory structure.
Most closely related to RIME, D-Mem~\cite{you2026dmem} adopts a dual-process
design that falls back to Full Deliberation over the interaction history when
compact memory is insufficient.
RIME also revisits the original dialogue, but retrieves only query-relevant
source turns together with local context rather than deliberating over the
broader history.

\section{Methodology}
\label{sec:method}
\subsection{Retrieval-based Memory Formation}
RIME introduces complementary innovations in both memory formation and memory use.

\noindent\textbf{Question-guided evidence recall.}
RIME maintains a fixed set of generic formation questions
\begin{equation}
\mathcal{Q}=\{q_r\}_{r=1}^{R},
\end{equation}
which are shared across all sessions and independent of downstream queries.
In our implementation, $R=6$, covering people and relationships, events,
activities, plans, states and preferences, and specific factual details.

Let $\phi(\cdot)$ denote the text embedding function and
\begin{equation}
s(a,b)
=
\frac{\phi(a)^{\top}\phi(b)}
{\|\phi(a)\|_2\|\phi(b)\|_2}
\end{equation}
denote cosine similarity.
For each formation question $q_r$, RIME retrieves the $k_f$ most relevant
dialogue turns from the current session $\mathcal{S}_t$:
\begin{equation}
\mathcal{R}_t^{(r)}
=
\operatorname*{arg\,topk}_{x\in\mathcal{S}_t}^{k_f}
s(q_r,x).
\end{equation}
The formation questions therefore act as retrieval cues rather than
summarization instructions: each question independently surfaces dialogue
evidence associated with a particular type of information before memory
construction.

We first merge the turns recalled by all formation questions:
\begin{equation}
\mathcal{R}_t
=
\bigcup_{r=1}^{R}\mathcal{R}_t^{(r)}.
\end{equation}
Because some recalled turns depend on nearby context, RIME performs two rounds of local context expansion:
\begin{equation}
\widetilde{\mathcal{R}}_t^{(\ell+1)}
=
\widetilde{\mathcal{R}}_t^{(\ell)}
\cup
\mathcal{N}_t\!\left(\widetilde{\mathcal{R}}_t^{(\ell)}\right),
\qquad
\ell=0,1,
\end{equation}
with $\widetilde{\mathcal{R}}_t^{(0)}=\mathcal{R}_t$, where $\mathcal{N}_t(\cdot)$ adds neighboring turns needed to resolve local dialogue dependencies. The final evidence set is
\begin{equation}
\mathcal{E}_t=\widetilde{\mathcal{R}}_t^{(2)}.
\end{equation}

\noindent\textbf{Historical memory grounding.}
For each formation question $q_r$, RIME combines the question with its recalled evidence $\mathcal{R}_t^{(r)}$ to construct a query for retrieving relevant historical memories:
\begin{equation}
z_t^{(r)}
=
\operatorname{Concat}
\left(
q_r,\mathcal{R}_t^{(r)}
\right),
\end{equation}
where the dialogue turns in $\mathcal{R}_t^{(r)}$ are concatenated in decreasing order of retrieval similarity. The corresponding historical memories are retrieved as
\begin{equation}
\mathcal{H}_t^{(r)}
=
\operatorname*{arg\,topk}_{m\in\mathcal{M}_{t-1}}^{k_h}
s\!\left(z_t^{(r)},m\right),
\qquad
\mathcal{H}_t
=
\bigcup_{r=1}^{R}\mathcal{H}_t^{(r)}.
\end{equation}

\noindent\textbf{Joint memory formation and reconciliation.}
Given the recalled current-session evidence $\mathcal{E}_t$ and the retrieved
historical memories $\mathcal{H}_t$, RIME performs fact extraction,
within-session deduplication, reference resolution, and historical
reconciliation jointly in a single LLM call:
\begin{equation}
\mathcal{O}_t
=
F_{\mathrm{mem}}
\left(
\mathcal{E}_t,\mathcal{H}_t
\right).
\end{equation}
Each predicted operation $o\in\mathcal{O}_t$ takes one of four forms:
\begin{equation}
\operatorname{op}(o)
\in
\{
\textsc{Add},
\textsc{Update},
\textsc{Delete},
\textsc{NoOp}
\}.
\end{equation}
\textsc{Add} creates a new memory, \textsc{Update} revises or augments an
existing memory describing the same underlying fact, \textsc{Delete} removes
a memory explicitly contradicted as erroneous, and \textsc{NoOp} is used when
the historical memory already captures the current evidence. Each memory also
retains its temporal information and dialogue provenance.

Finally, the predicted operations are applied to obtain the updated memory
bank:
\begin{equation}
\mathcal{M}_t
=
\operatorname{Apply}
\left(
\mathcal{M}_{t-1},\mathcal{O}_t
\right).
\end{equation}

\subsection{Selective Memory Use}
\label{sec:memory_use}

Given a user query $q_i^{\mathrm{usr}}$, let
$\tilde{y}_i=F_{\mathrm{ans}}(q_i^{\mathrm{usr}},\mathcal{M}_i)$ denote the
initial answer from the retrieved semantic memories $\mathcal{M}_i$.
If $\tilde{y}_i=\texttt{NONE}$, RIME retrieves relevant source turns using
TF--IDF~\cite{salton1988term}:
\begin{equation}
\mathcal{R}_i^{\mathrm{src}}
=
\operatorname*{arg\,topk}_{x\in\bigcup_{t=1}^{T}\mathcal{S}_t}^{k_s}
\psi(q_i^{\mathrm{usr}})^\top\psi(x),
\quad
\mathcal{E}_i^{\mathrm{src}}
=
\mathcal{R}_i^{\mathrm{src}}
\cup
\operatorname{Ctx}_{w}(\mathcal{R}_i^{\mathrm{src}}),
\end{equation}
where $\operatorname{Ctx}_{w}(\cdot)$ adds up to $w$ preceding and following
turns within the same session. The final prediction is
\begin{equation}
\hat{y}_i=
\begin{cases}
\tilde{y}_i, & \tilde{y}_i \neq \texttt{NONE}, \\[2pt]
F_{\mathrm{ans}}(q_i^{\mathrm{usr}},\mathcal{E}_i^{\mathrm{src}}),
& \tilde{y}_i = \texttt{NONE}.
\end{cases}
\end{equation}

\section{Experiments}
\label{sec:experiments}

\subsection{Experimental Setup}
\label{sec:setup}

\noindent\textbf{Dataset.}
We evaluate on LoCoMo~\cite{maharana2024locomo}, a benchmark for very long-term conversational memory. Following the paper-comparable setting, we use 1,540 questions from four categories: single-hop, multi-hop, temporal, and open-domain, excluding the adversarial category.
\noindent\textbf{Baselines.}
We compare RIME with RAG~\cite{lewis2020rag}, Mem0~\cite{chhikara2025mem0}, A-MEM~\cite{xu2026mem}, Nemori~\cite{ma2026deserves}, and D-Mem~\cite{you2026dmem}, covering retrieval-based, structured, adaptive, and deliberative long-term memory systems.
\noindent\textbf{Models.}
We evaluate all methods with two LLM backbones, Qwen3-235B-A22B and GPT-5.6 Sol. For each setting, the same backbone is used for memory construction and answer generation, and the corresponding model is also used as the LLM judge.
For answer generation and evaluation, we follow the prompting protocol used by D-Mem~\cite{you2026dmem} and Nemori~\cite{ma2026deserves}, including the answer and LLM-judge prompts.
\noindent\textbf{Memory and Retrieval Configuration.}
RIME uses six generic formation queries covering person, event, activity, plan, state, and specific detail. 
For each query, we retrieve the top 10 dialogue turns and top 5 relevant historical memories. 
At inference, we retrieve the top 30 semantic memories using OpenAI's \texttt{text-embedding-3-small}. 
If the initial answer is \texttt{NONE}, TF--IDF retrieval selects the top 20 source-dialogue turns, each expanded with a context window of 15 turns. No raw dialogue is provided during the initial memory-based answer. 
Unless otherwise specified, all baselines use the same retrieval settings; A-MEM retains its original \texttt{all-MiniLM-L6-v2} embedding model following its official configuration.
The source code, including the exact prompts
and evaluation configuration, is available at \url{https://anonymous.4open.science/r/RIME_official-4BC7}.

\subsection{Main Results}
\label{sec:main_results}

Table~\ref{tab:main_results} compares RIME with representative long-term
memory systems under two LLM backbones. RIME consistently achieves the best
performance across all three quality metrics with both Qwen3-235B-A22B and
GPT-5.6 Sol. Notably, these gains are achieved with substantially lower
query-time LLM cost than deliberative memory baselines such as A-MEM and
Nemori, demonstrating that selective access to source dialogue provides an
effective alternative to expensive deliberation over long histories.

\begin{table}[t]
\centering
\caption{
Main results on the 1,540 non-adversarial LoCoMo questions.
Higher is better for Judge, F1, and BLEU-1; lower is better for inference tokens.
$^\dagger$D-Mem results are directly reported from the original paper using
Qwen3-235B-Instruct, as its implementation is not publicly available.
}
\label{tab:main_results}
\small
\setlength{\tabcolsep}{4pt}
\renewcommand{\arraystretch}{1.08}
\begin{tabular*}{\linewidth}{@{\extracolsep{\fill}}lrrrr@{}}
\toprule
Method & Judge$\uparrow$ & F1$\uparrow$ & BLEU-1$\uparrow$ & Tokens$\downarrow$ \\
\midrule
\multicolumn{5}{l}{\textit{Qwen3-235B-A22B}} \\
RAG    & 31.36 & 20.87 & 17.46 & 3.64K \\
Mem0   & 53.12 & 35.76 & 30.49 & 2.20K \\
A-MEM  & 68.77 & 44.12 & 37.86 & 63.19K \\
Nemori & 77.34 & 46.01 & 39.88 & 20.92K \\
D-Mem$^\dagger$
       & 78.60 & 51.00 & 42.60 & 15.57K \\
RIME   & \textbf{81.82} & \textbf{51.46} & \textbf{44.22} & 5.11K \\
\midrule
\multicolumn{5}{l}{\textit{GPT-5.6 Sol}} \\
RAG    & 40.00 & 23.94 & 20.29 & 3.31K \\
Mem0   & 70.52 & 46.45 & 39.59 & 2.24K \\
A-MEM  & 80.78 & 53.77 & 46.54 & 52.14K \\
Nemori & 83.05 & 52.64 & 45.88 & 17.74K \\
RIME   & \textbf{84.29} & \textbf{56.41} & \textbf{48.85} &5.21K \\
\bottomrule
\end{tabular*}
\end{table}

\subsection{Ablation and Analysis}
\label{sec:analysis}
\noindent\textbf{Effect of question-guided memory formation.}
To isolate the effect of question-guided retrieval, we compare three formation
strategies with the same memory-use stage.
\textit{Direct Compression} uses the Mem0 extraction prompt over the full
session, while \textit{Question-Prompted Compression} additionally uses the
six formation dimensions as extraction instructions without question-specific
retrieval. RIME instead uses these questions as retrieval cues before memory
consolidation.
As shown in Table~\ref{tab:formation_analysis}, adding the six dimensions to
the compression prompt yields no consistent gain over direct compression,
whereas RIME improves all three metrics.
This suggests that question-guided memory formation is effective.

\noindent\textbf{Effect of selective source retrieval.}
We further examine whether access to the source dialogue is necessary when
semantic memory is insufficient. Disabling selective source retrieval causes
a substantial performance drop on both backbones, with the Judge score
decreasing by 10.59 points on Qwen3-235B-A22B and 3.32 points on GPT-5.6 Sol.
Source retrieval is triggered more frequently with Qwen than with GPT-5.6
(16.8\% vs.\ 6.0\%), consistent with the larger degradation observed when
this mechanism is removed.

\begin{table}[t]
\centering
\caption{
Effect of memory formation strategies on LoCoMo using Qwen3-235B-A22B
($n=1{,}540$). All variants use the same selective memory-use procedure.
}
\label{tab:formation_analysis}
\small
\setlength{\tabcolsep}{5pt}
\renewcommand{\arraystretch}{1.08}
\begin{tabular*}{\linewidth}{@{\extracolsep{\fill}}lccc@{}}
\toprule
Formation Strategy & Judge$\uparrow$ & F1$\uparrow$ & BLEU-1$\uparrow$ \\
\midrule
Direct Compression
& 71.62 & 49.26 & 42.32 \\
Question-Prompted Compression
& 72.66 & 48.87 & 42.11 \\
RIME
& \textbf{81.82} & \textbf{51.46} & \textbf{44.22} \\
\bottomrule
\end{tabular*}
\end{table}

\begin{table}[t]
\centering
\caption{Ablation of selective source retrieval on LoCoMo.}
\label{tab:ablation_selective}
\small
\setlength{\tabcolsep}{5pt}
\begin{tabular*}{\linewidth}{@{\extracolsep{\fill}}lccc@{}}
\toprule
Method & Judge$\uparrow$ & F1$\uparrow$ & BLEU-1$\uparrow$ \\
\midrule
\multicolumn{4}{l}{\textit{Qwen3-235B-A22B}} \\
RIME w/o selective retrieval & 71.23 & 44.73 & 38.41 \\
RIME                         & \textbf{81.82} & \textbf{51.46} & \textbf{44.22} \\
\midrule
\multicolumn{4}{l}{\textit{GPT-5.6 Sol}} \\
RIME w/o selective retrieval & 80.97 & 54.27 & 46.96 \\
RIME                         & \textbf{84.29} & \textbf{56.41} & \textbf{48.85} \\
\bottomrule
\end{tabular*}
\end{table}

\section{Conclusion}
\label{sec:conclusion}

We presented RIME, a retrieval-induced framework for long-term dialogue memory.
RIME uses generic questions as retrieval cues to identify relevant evidence
before memory consolidation and selectively returns to the source dialogue
when compressed memory is insufficient.
Experiments on LoCoMo with Qwen3-235B-A22B and GPT-5.6 Sol show that RIME
achieves the best performance across LLM-Judge, F1, and BLEU-1 among the
compared methods, while using substantially fewer query-time LLM tokens than
deliberative memory baselines.
Further analysis confirms the effectiveness of both question-guided memory
formation and selective source retrieval for robust and efficient long-term memory.


\bibliographystyle{IEEEbib}
\bibliography{strings,refs}

@inproceedings{ma2026deserves,
  title={What deserves memory: Adaptive memory distillation for llm agents},
  author={Ma, Wenquan and Nan, Jiayan and Wu, Wenlong},
  booktitle={Proceedings of the 64th Annual Meeting of the Association for Computational Linguistics (Volume 1: Long Papers)},
  pages={34789--34812},
  year={2026}
}

@inproceedings{park2023generative,
  title={Generative agents: Interactive simulacra of human behavior},
  author={Park, Joon Sung and O'Brien, Joseph and Cai, Carrie Jun and Morris, Meredith Ringel and Liang, Percy and Bernstein, Michael S},
  booktitle={Proceedings of the 36th annual acm symposium on user interface software and technology},
  pages={1--22},
  year={2023}
}

@inproceedings{zhong2024memorybank,
  title={Memorybank: Enhancing large language models with long-term memory},
  author={Zhong, Wanjun and Guo, Lianghong and Gao, Qiqi and Ye, He and Wang, Yanlin},
  booktitle={Proceedings of the AAAI conference on artificial intelligence},
  volume={38},
  number={17},
  pages={19724--19731},
  year={2024}
}

@article{packer2023memgpt,
  title={Memgpt: Towards llms as operating systems},
  author={Packer, Charles and Wooders, Sarah and Lin, Kevin and Fang, Vivian and Patil, Shishir G and Stoica, Ion and Gonzalez, Joseph E},
  journal={arXiv preprint arXiv:2310.08560},
  year={2023}
}

@article{lewis2020rag,
  title={Retrieval-augmented generation for knowledge-intensive nlp tasks},
  author={Lewis, Patrick and Perez, Ethan and Piktus, Aleksandra and Petroni, Fabio and Karpukhin, Vladimir and Goyal, Naman and K{\"u}ttler, Heinrich and Lewis, Mike and Yih, Wen-tau and Rockt{\"a}schel, Tim and others},
  journal={Advances in neural information processing systems},
  volume={33},
  pages={9459--9474},
  year={2020}
}

@inproceedings{chhikara2025mem0,
  author       = {Prateek Chhikara and
                  Dev Khant and
                  Saket Aryan and
                  Taranjeet Singh and
                  Deshraj Yadav},
  title        = {Mem0: Building Production-Ready {AI} Agents with Scalable Long-Term
                  Memory},
  booktitle    = {{ECAI}},
  series       = {Frontiers in Artificial Intelligence and Applications},
  pages        = {2993--3000},
  publisher    = {{IOS} Press},
  year         = {2025}
}

@article{xu2026mem,
  title={A-mem: Agentic memory for llm agents},
  author={Xu, Wujiang and Liang, Zujie and Mei, Kai and Gao, Hang and Tan, Juntao and Zhang, Yongfeng},
  journal={Advances in Neural Information Processing Systems},
  volume={38},
  pages={17577--17604},
  year={2026}
}

@article{simplemem2026,
  title={Simplemem: Efficient lifelong memory for llm agents},
  author={Liu, Jiaqi and Su, Yaofeng and Xia, Peng and Han, Siwei and Zheng, Zeyu and Xie, Cihang and Ding, Mingyu and Yao, Huaxiu},
  journal={arXiv preprint arXiv:2601.02553},
  year={2026}
}

@article{you2026dmem,
  title={D-Mem: A Dual-Process Memory System for LLM Agents},
  author={You, Zhixing and Yuan, Jiachen and Cai, Jason},
  journal={arXiv preprint arXiv:2603.18631},
  year={2026}
}

@inproceedings{maharana2024locomo,
  title={Evaluating very long-term conversational memory of llm agents},
  author={Maharana, Adyasha and Lee, Dong-Ho and Tulyakov, Sergey and Bansal, Mohit and Barbieri, Francesco and Fang, Yuwei},
  booktitle={Proceedings of the 62nd Annual Meeting of the Association for Computational Linguistics (Volume 1: Long Papers)},
  pages={13851--13870},
  year={2024}
}

@article{wang2026sage,
  title={SAGE: A Novelty Gate for Efficient Memory Evolution in Agentic LLMs},
  author={Wang, Sijia and Brahma, Dhanajit and Henao, Ricardo},
  journal={arXiv preprint arXiv:2605.30711},
  year={2026}
}

@inproceedings{wei2026fademem,
  title={Fademem: Biologically-inspired forgetting for efficient agent memory},
  author={Wei, Lei and Dong, Xu and Peng, Xiao and Xie, Niantao and Wang, Bin},
  booktitle={ICASSP 2026-2026 IEEE International Conference on Acoustics, Speech and Signal Processing (ICASSP)},
  pages={4011--4015},
  year={2026},
  organization={IEEE}
}

@article{yang2025qwen3,
  title={Qwen3 technical report},
  author={Yang, An and Li, Anfeng and Yang, Baosong and Zhang, Beichen and Hui, Binyuan and Zheng, Bo and Yu, Bowen and Gao, Chang and Huang, Chengen and Lv, Chenxu and others},
  journal={arXiv preprint arXiv:2505.09388},
  year={2025}
}

@misc{openai2026gpt56,
  title        = {GPT-5.6 System Card},
  author       = {{OpenAI}},
  year         = {2026},
  month        = jul,
  howpublished = {\url{https://deploymentsafety.openai.com/gpt-5-6}},
}

@article{salton1988term,
  title={Term-weighting approaches in automatic text retrieval},
  author={Salton, Gerard and Buckley, Christopher},
  journal={Information processing \& management},
  volume={24},
  number={5},
  pages={513--523},
  year={1988},
  publisher={Elsevier}
}

@inproceedings{fang2026lightmem,
  title={Lightmem: Lightweight and efficient memory-augmented generation},
  author={Fang, Jizhan and Deng, Xinle and Xu, Haoming and Jiang, Ziyan and Tang, Yuqi and Xu, Ziwen and Deng, Shumin and Yao, Yunzhi and Wang, Mengru and Qiao, Shuofei and others},
  booktitle={International Conference on Learning Representations},
  volume={2026},
  pages={98706--98729},
  year={2026}
}

@inproceedings{latimer2026hindsight,
  title={HINDSIGHT: Structured Agent Memory that Retains, Recalls, and Reflects},
  author={Latimer, Christopher and Boschi, Nicol{\'o} and Neeser, Andrew and Bartholomew, Chris and Srivastava, Gaurav and Wang, Xuan and Ramakrishnan, Naren},
  booktitle={Proceedings of the 64th Annual Meeting of the Association for Computational Linguistics (Volume 3: System Demonstrations)},
  pages={275--285},
  year={2026}
}

@article{song2026retrieval,
  title={Retrieval-Driven Memory Reconsolidation for Long-Term LLM Agents},
  author={Song, Yuanyi and Wang, Yukai and Ma, Xinbei and Fu, Zhihui and Lin, Jianghao and Liu, Weiwen and Wang, Jun and Deng, Huarong and Yu, Yong and Zhang, Weinan},
  journal={arXiv preprint arXiv:2609.16053},
  year={2026}
}

\end{document}